 \documentclass[pmlr,onecolumn]{jmlr} 

\usepackage{mathtools}
\usepackage{multirow}
\newcommand\myeq{\stackrel{\mathclap{\normalfont\mbox{def}}}{=}}
\DeclareMathOperator*{\argmin}{arg\,min}
\DeclareMathOperator*{\argmax}{arg\,max}

\usepackage{boldline, makecell}
\usepackage{todonotes}

\usepackage{booktabs}
\usepackage[load-configurations=version-1]{siunitx} 

\theorembodyfont{\upshape}
\theoremheaderfont{\scshape}
\theorempostheader{:}
\theoremsep{\newline}

\jmlrvolume{}
\jmlryear{2020}
\jmlrsubmitted{}
\jmlrpublished{}
\jmlrworkshop{16-811 Math Fundamentals Project} 

\usepackage{bbm}

\title[Learning without Pointillistic Labels]{The Word is Mightier than the Label \\ Learning without Pointillistic Labels using Data Programming} 

  \author{
  \Name{Chufan Gao} \Email{chufang@andrew.cmu.edu}\\
  \Name{Mononito Goswami} \Email{mgoswami@andrew.cmu.edu}\\
  \addr Robotics Institute, Carnegie Mellon University
  }

\begin{document}

\maketitle

\begin{abstract}
Most advanced supervised Machine Learning (ML) models rely on vast amounts of point-by-point labelled training examples. Hand-labelling vast amounts of data may be tedious, expensive, and error-prone. Recently, some studies have explored the use of diverse sources of weak supervision to produce competitive end model classifiers. In this paper, we survey recent work on weak supervision, and in particular, we investigate the Data Programming (DP) framework \citep{ratner2016data}. Taking a set of potentially noisy heuristics as input, DP assigns denoised probabilistic labels to each data point in a dataset using a probabilistic graphical model of heuristics. We analyze the math fundamentals behind DP and demonstrate the power of it by applying it on two real-world text classification tasks. Furthermore, we compare DP with pointillistic active and semi-supervised learning techniques traditionally applied in data-sparse settings. 
\end{abstract}
\begin{keywords}
Data programming, weak supervision, active learning, semi-supervised learning, text classification
\end{keywords} 
\section{Introduction}

Advances in supervised machine learning (ML) have enabled ML models to perform at par with humans in several classification and decision making settings. However, most advanced ML models such as deep neural networks (NNs) rely on vast quantities of pointillistically labelled training data. While unlabelled raw data is abundant, hand-labelling large quantities of data is tedious, expensive, time intensive and prone to error. For instance, \citep{hannun2019cardiologist} recently developed a model to diagnose irregular heartbeats (arrhythmia) from ECG signals, better than individual cardiologists\footnote{\url{https://stanfordmlgroup.github.io/projects/ecg/}}. However, their Convolutional Neural Network (CNN) was trained on as many as $64,121$ ECG records from $29,163$ patients, each hand-labelled by an expert cardiologist. Another example is the prostate, lung, colorectal and ovarian (PLCO) cancer screening trial of the National Cancer Institute, a dataset of several thousand chest radiographs collected over 13 years \citep{team2000prostate}. The healthcare domain is replete with ML models achieving amazing results, limited only by availability of large, cleaned and annotated datasets. Additionally, in many other cases, obtaining millions of unlabelled images from cameras, videos, and vast quantities of textual data from the web is straightforward -- the burden lies in annotation.

Many ML techniques have been devised to train competitive models by actively and intelligently sampling training data \citep{settles2009active}, combining both unlabelled and labelled data \citep{van2020survey} or harnessing crowd-sourced labels \citep{gao2011harnessing}. In this work, we explore the use of diverse sources of weak supervision to train competitive end model classifiers. We use the recently introduced \textit{data programming} framework to intelligently combine these weak supervision sources to probabilistically label data and train competitive downstream classifiers without access to ground truth labels \citep{ratner2016data}. 

Our major contribution in this study is to come up with simple ways to automatically acquire weak supervision sources for text classification problems with minimal human effort. Furthermore, we compare end model classifiers obtained using weak supervision, full supervision, active learning and semi-supervised learning. Our experiments on the widely used IMDb review sentiment classification dataset \citep{maas-EtAl:2011:ACL-HLT2011} and the 20 Newsgroups dataset \citep{Lang95}, reveal that weak supervision performs at par with its fully supervised counterpart, and outperforms alternatives such as active and semi-supervision. Our results are also consistent with prior work which show that \textit{keywords} (unigrams) are excellent sources of weak supervision \citep{boecking2019pairwise, ratner2016data}.  

The rest of the paper is organised as follows. In section~\ref{sec:priorwork} we do a brief literature survey of current approaches to ML data-sparse settings. Section~\ref{dataprogramming} briefly overviews the data programming label framework and label model. In section~\ref{experiments}, we describe our experimental setup, the acquisition of weak supervision sources and datasets. We conclude the paper with discussion of our results in Section~\ref{resultsanddiscussion}.




\label{sec:intro}

\section{Prior Work and Alternative Approaches}

\label{sec:priorwork}
In this section, we will overview 2 baseline techniques commonly used in machine learning with less labels.

\subsection{Active Learning}
\begin{algorithm}[h!]
\SetAlgoLined
Let $\mathcal{S} = \{(x_i, y_i)\}_{i = 1}^{n}, \ (x, y) \sim \mathcal{X} \times \mathcal{Y} \cup \{-1\}$ represent the training dataset, where $-1$ represents the unlabelled data points. We can then partition $\mathcal{S}$ into disjoint labelled $\mathcal{S}_{l} = \{(x, y)\}$ such that $y \in \mathcal{Y}$ and unlabelled subsets $\mathcal{S}_{u} = \{(x, y)\}$ such that $y = -1$. Let $\mathbb{T}$ represent the predefined number of active learning iterations. Our goal is to find a hypothesis $h_\mathbb{T} \in \mathcal{H}$ which minimizes empirical risk under an arbitrary loss function $l$.\\

\For{$t\gets0$ in $\mathbb{T}$}{
    $h_t = \underset{h_t \in \mathcal{H}}{\argmin} \, \mathbb{E}_{(h_t(x), y) \in \mathcal{S}}[l(x, y)]$\\
    $\mathcal{P} = h_t.\texttt{predict\_proba}(\mathcal{S}_{u})$ // \texttt{predict\_proba} returns probabilistic labels $(p(y = 1), \ p(y = 0))$ for each $(x, y) \in \mathcal{S}_u$ \\
    $\mathcal{X}_{query}$ = \texttt{acquisition\_function($\mathcal{P}$)} // $\mathcal{X}_{query}$ is a set of unlabelled data points. 
    $\mathcal{Y}_{query}$ = \texttt{get\_labels}($\mathcal{X}_{query}$) // \texttt{get\_labels} is a function to acquire labels from the oracle \\
    $\mathcal{S}_l := \mathcal{S}_l \cup \{\mathcal{X}_{query} , \mathcal{Y}_{query} \}$  \\
    $\mathcal{S}_u := \mathcal{S}_u - \{\mathcal{X}_{query} , \mathcal{Y}_{query} \}$  \\
}
\caption{Active learning pseudocode}
\label{alg:active}
\end{algorithm}

The active learning process is straightforward--the dataset is split into two subsets, labeled and unlabeled. The algorithm is trained on the labeled dataset, where it predicts labels on the unlabeled dataset. Depending on the confidence on the predicted labels, queries are made for the expert to label more data. The new labeled data is then added to the existing labeled dataset.
Algorithm \ref{alg:active} shows the complete psuedocode.

Additionally, there are multiple methods in considering the ``confidence" on predicted labels. One of which is entropy based, where we measure the predicted labels based on Shannon Entropy $H(x) = -\sum P(x) log(P(x))$, where we ask the oracle to label high entropy label predictions. The other is simple taking the prediction probability of the prediction labels and asking the oracle to label low probability label predictions.

\subsection{Semi-supervised learning}
\begin{algorithm}[h!]
\SetAlgoLined
Let $\mathcal{S} = \{(x_i, y_i)\}_{i = 1}^{n}, \ (x, y) \sim \mathcal{X} \times \mathcal{Y} \cup \{-1\}$ represent the training dataset, where $-1$ represents the unlabelled data points. We can then partition $\mathcal{S}$ into disjoint labelled $\mathcal{S}_{l} = \{(x, y)\}$ such that $y \in \mathcal{Y}$ and unlabelled subsets $\mathcal{S}_{u} = \{(x, y)\}$ such that $y = -1$. Let $\mathbb{T}$ represent the predefined number of active learning iterations. Our goal is to find a hypothesis $h_\mathbb{T} \in \mathcal{H}$ which minimizes empirical risk under an arbitrary loss function $l$. Let $\mathcal{C}$ be the cutoff threshold.\\

\For{$t\gets0$ in $\mathbb{T}$}{
    $h_t = \underset{h_t \in \mathcal{H}}{\argmin} \, \mathbb{E}_{(h_t(x), y) \in \mathcal{S}}[l(x, y)]$\\
    $\mathcal{P} = h_t.\texttt{predict\_proba}(\mathcal{S}_{u})$ // \texttt{predict\_proba} returns probabilistic labels $(p(y = 1), \ p(y = 0))$ for each $(x, y) \in \mathcal{S}_u$ \\
    $\mathcal{X}_{confident} = \mathcal{S}_u[\texttt{argmax}(\mathcal{P}) > \mathcal{C}]$ is the data points corresponding confident predictions. \\ 
    $\mathcal{Y}_{confident}$ = $\mathcal{P}[\texttt{argmax}(\mathcal{P}) > \mathcal{C}]$ //  the predicted labels \\
    $\mathcal{S}_l := \mathcal{S}_l \cup \{\mathcal{X}_{confident} , \mathcal{Y}_{confident} \}$  \\
    $\mathcal{S}_u := \mathcal{S}_u - \{\mathcal{X}_{confident} , \mathcal{Y}_{confident} \}$  \\
}
\caption{Semi-supervised learning pseudocode}
\label{alg:semi}
\end{algorithm}

Semi-supervised learning is very similar to Active Learning, but with one key difference. The problem setup is the same, where we split the dataset into unlabeled and labeled subsets. For each iteration of semi-supervised learning, the model is trained on the labeled dataset, then is used to predict labels in the unlabeled dataset. The label predictions with the highest confidence is added to the labeled dataset with the predicted labels (without any input from an oracle). Algorithm \ref{alg:semi} shows the complete psuedocode.

\section{Data Programming}
\label{dataprogramming}
\begin{figure}[h!]
    \centering
    \includegraphics[width=\linewidth]{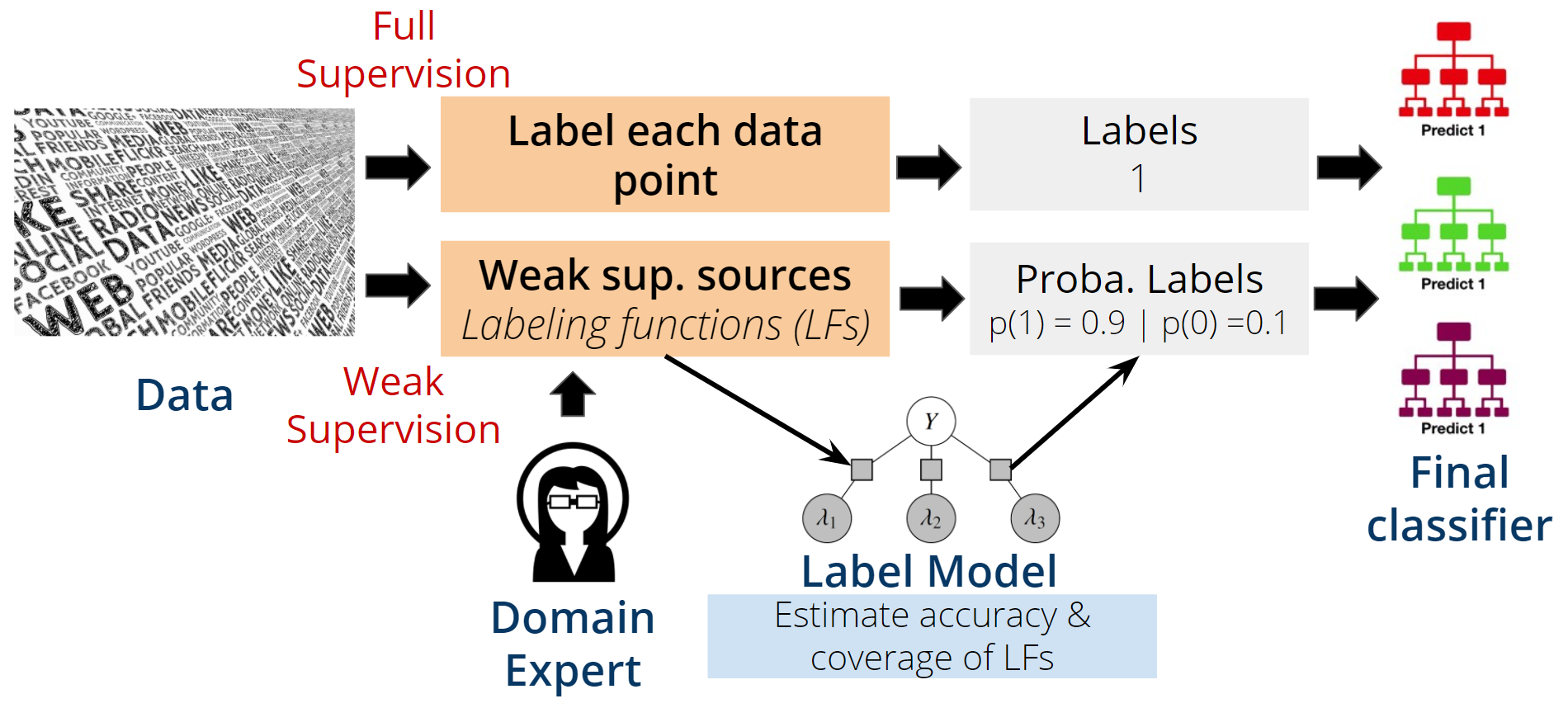}
    \caption{Flowchart of the Data Programming process compared to traditional supervised machine learning. The orange boxes indicate the effort required by the users (annotators). Instead of having to label everything by hand, the work lies only in defining some labeling functions, which is much more efficient.
}
    \label{fig:flowchart}
\end{figure}
In this section, we shall provide a brief overview of the data programming model. Figure \ref{fig:flowchart}) shows the overall advantages of this weak supervision model compared to full supervision. For the remainder of the section, we will explain the mathematical background behind data programming. 

Consider a binary classification task in which we have a distribution $\pi$ over object-class pairs $(x, y) \in \mathcal{X} \times \{0, 1\}$. To specify a data programming model, users define a number of \textit{labelling functions} (LFs), $\lambda \myeq \mathcal{X} \mapsto \{-1, 0, 1\}$, which encode domain knowledge to noisily label subsets of the data. 

Given $n$ i.i.d training object $\mathcal{S}_x = \{x_1, x_2, \dots, x_n\} \in \mathcal{X}$ and a set of $m$ labelling functions $\Lambda = \{\lambda_1, \dots, \lambda_m\}$, data programming infers probabilistic labels $\mathcal{Y} \in [0, 1]^n$ for all training objects in $\mathcal{S}_x$. We can then use these probabilistic labels to train a downstream classifier via \textit{Empirical Risk Minimization} (ERM) under a noise-aware loss. Given training data of the form $\mathcal{S} = ((x_1, y_1), \dots, (x_n, y_n)), \ x_i \in \mathcal{X}, \ y_i \in \{0, 1\}$, the ERM algorithm outputs the hypothesis $h \in \mathcal{H}$ which minimizes empirical risk:
\begin{equation*}
    \hat h = \underset{h \in \mathcal{H}}{\argmin} \frac{1}{|S|} \sum_{x, y \ \in \ S} l(h(x), \ y)
\end{equation*}
where $l$ can be any loss function. Due to the probabilistic nature of the labels $\mathcal{Y}$ inferred by data programming label model, we can instead minimize a noise-aware empirical risk, which weighs the loss incurred by training objects in proportion to the label model's confidence on its own label assignment to the object. In other words, a noise-aware loss may give more weight to those training objects which the label model is confident about (having higher values of $y \in \mathcal{Y}$). 

\subsection{Label Model}
Now we will describe the data programming label model which assumes that the \textit{LFs $\Lambda$ label independently given the true latent class labels $Y$.} Under this model, let us assume that each LF $\lambda_i$ labels an object with probability $\beta_i$, and with probability $\alpha_i$ labels the object correctly. We also assume that the positive and negative class are equally prevalent, \textit{i.e.} each class has a probability of $\frac{1}{2}$. This model has the following distribution:
\begin{equation}
    \mu_{\alpha, \beta}(\Lambda, Y) = \frac{1}{2} \prod_{i = 1}^m \left( \beta_i \alpha_i \mathbbm{1}_{\Lambda_i = Y} + \beta_i (1 - \alpha_i) \mathbbm{1}_{\Lambda_i = -Y} + (1 - \beta_i) \mathbbm{1}_{\Lambda_i = 0} \right)
    \label{labelmodel}
\end{equation}
where $\Lambda = \{-1, 0, 1\}^m$ are the labels output by the LFs and $Y \in \{-1, 1\}$ are the latent predicted class labels. When we allow the parameters $\alpha \in \mathbb{R}^m$ and $\beta \in \mathbb{R}^m$ to vary, Eqn.~\ref{labelmodel} defines a family of generative models. 

Our goal is to learn the parameters $\alpha$ and $\beta$ which are the most consistent with our observations \textit{i.e.} the unlabelled training data $\mathcal{S}_x$, using maximum likelihood estimation. For a particular training set $\mathcal{S}_x \in \mathcal{X}$, we solve the following problem: 
\begin{equation}
    (\hat{\alpha}, \hat{\beta}) = \underset{\alpha, \beta}{\argmax} \sum_{x \in \mathcal{S}_x} \log \mathbb{P}_{(\Lambda, Y) \sim \mu_{\alpha, \beta}} (\Lambda = \lambda(x))
    \label{mlealphabeta}
\end{equation}
Hence, our goal is to maximize the likelihood that the observed labels $(\lambda(x))$ occur under the generative model defined in Eqn.~\ref{labelmodel}. 

\subsection{From Noisy LF votes to Probabilistic Labels}
The noisy labels $\Lambda$ from the LFs can be aggregated using an arbitrary rule into ``denoised'' probabilistic labels. The quality of the final labels, in addition to the LFs, also depends on the aggregation rule \citep{li2015theoretical}. \textit{Majority voting} is a natural choice for an aggregation rule. While majority voting often works well in practice, it weighs each LF the same. However, in practice we may want to weigh more accurate LFs more than the less accurate ones. Hence, data programming uses the weighted majority vote, weighing each vote of a LF $\lambda_i$ by its accuracy $\alpha_i$. 

Before formally describing the weighted aggregation rule, we shall assume the following notation. Let $[m] = \{1, 2, ... m\}$ and $\hat{y_i}$ denote the predicted label of $x_i \in \mathcal{S}_x$. Also let $\mathbb{V}_{n \times m} \in \{-1, 0, 1\}^{n \times m}$ denote the \textit{vote matrix}, such that $v_{ij} \in \{-1, 0, 1\}$ corresponds to the vote of $\lambda_j \in \Lambda$ on $x_i \in \mathcal{S}_x$. Then weighted majority aggregation rule can be formally defined as:
\begin{equation}
    \hat{y_i} = \text{sign}(\sum_{j = 1}^m w_j\mathbb{V}_{ij})    
    \label{eqn:weightedmajorityvoting}
\end{equation}

We can derive the majority vote aggregation rule by setting $w_j = 1$. 

In addition to the predicted label $\hat y_i$, data programming also evaluates probabilistic labels given by the pair $(\mathbb{P}(\hat{y_i} = 1), \mathbb{P}(\hat{y_i} = -1))$. These probabilitis are computed using the softmax function as follows: 
\begin{align}
    \mathbb{P}(\hat{y_i} = 1) & = \frac{\exp{(\sum_{j = 1}^m w_j \mathbbm{1}_{\mathbb{V}_{ij} = 1 }})}{\exp{(\sum_{j = 1}^m w_j \mathbbm{1}_{\mathbb{V}_{ij} = 1 }}) + \exp{(\sum_{j = 1}^m w_j \mathbbm{1}_{\mathbb{V}_{ij} = -1}})} \\
    \mathbb{P}(\hat{y_i} = -1) & = 1 - \mathbb{P}(\hat{y_i} = 1)
    \label{eqn:probabilisticlabels}
\end{align}
where $w_j = \frac{\hat \alpha_j}{\sum_{j = 1}^m \hat \alpha_j}$ and $\hat \alpha_j$ is derived from the label model in Eqn.~\ref{mlealphabeta}.

\section{Experiments}
\label{experiments}
In order to evaluate the efficacy of data programming, we compared it with four baselines, namely, fully supervised classification (FS), active learning (AL), semi-supervised learning (SSL) and majority vote (Maj.) label model. For all our experiments, we used the random forest implementation of \texttt{scikit-learn}\footnote{\url{https://scikit-learn.org/stable/modules/generated/sklearn.ensemble.RandomForestClassifier.html}} comprising of $500$ decision tree estimators. We chose random forest over other classifiers because of its ability to learn complex decision boundaries without overfitting to training data \citep{goswami2020discriminating, goswami2020towards}. We used document embeddings from a pre-trained BERT model \citep{reimers2019sentencebert}\footnote{Specifically, we used the \texttt{sentence-transformers} (\url{https://www.sbert.net/}) Python package.} as featurization for our classifiers. 

\begin{table}[!tbp]
    \centering
    \resizebox{0.5\textwidth}{!}{
    \begin{tabular}{c c c}
         \textbf{Algorithm} & \textbf{Hyper-parameter} & \textbf{Value} \\ \hline
         \multirow{2}{*}{Random forests} & \# estimators & 500 \\
         & Min. samples in a leaf & 5 \\ \hline
         \multirow{3}{*}{Active learning} & Seed samples & 20 \\
         & Query size & 1 \\
         & \# iterations & 1000 \\ \hline
         \multirow{3}{*}{Semi-sup. learning} & Seed samples & 1000 \\
         & Batch size & 10 \\
         & \# iterations & 25 \\ \hline 
         \multirow{2}{*}{Weak supervision} & Optimizer & SGD \\
         & Learning rate & 0.01 \\ \hline
    \end{tabular}}
    \caption{Hyper-parameters of various models.}
    \label{tab:hyperparams}
\end{table}

We implemented two variants of active learning with uncertainty sampling, namely \textit{entropy}-based and \textit{least confidence} \citep{settles2009active}. Both the implementations were identical\footnote{For binary classification both variants are equivalent.} except the uncertainty measure used to query the most informative samples. While the former queried the instance having the highest \textit{entropy} \citep{10.1145/584091.584093}, the latter queried the instance whose best labelling is the least confident \textit{i.e.} choosing the instance with posterior closest to $0.5$ for binary classification. All our active learning experiments started with a randomly chosen class-balanced seed set of $20$ labelled objects. In each active learning iteration, the object with the highest entropy or least confidence was chosen to be labelled by the expert, for a total of $1000$ active learning iterations. 

On the other hand, our semi-supervised model used the self-labelling strategy \citep{van2020survey}. For our semi-supervised model we started with a randomly chosen class-balanced seed set of $1000$ labelled objects. In each semi-supervision iteration, we added a batch of $10$ most confidently labelled objects to the training set. We repeated this procedure for a total of $25$ semi-supervision iterations. The hyper-parameters of all our models are listed in Tab.~\ref{tab:hyperparams}. 

Furthermore, recall that our label model returns probabilistic labels of the form $(\mathbb{P}(\hat{y_i} = 1), \mathbb{P}(\hat{y_i} = -1))$. We can then formulate the confidence $\mathcal{C}_i \in [0, 1]$ of the label model on its label prediction on $x_i$ as: 
\begin{equation}
    \mathcal{C}_i = 2 \times \left(\max(\mathbb{P}(\hat{y_i} = 1), \mathbb{P}(\hat{y_i} = -1)) - \frac{1}{2} \right)
\end{equation}
where $\hat y_i$ is the label predicted by the label model. Using these confidence scores, we can weigh confidently labelled objects more than the uncertain ones. Since we could not introduce sample weights into \texttt{scikit-learn}'s random forest implementation, we instead only trained our weakly supervised model on samples with a confidence score greater than $\mathcal{C} = 0.6$.

\subsection{Datasets}
In this study, we use two datasets, the IMDb Large Movie Reviews (\texttt{IMDb})\footnote{\url{https://ai.stanford.edu/~amaas/data/sentiment/}} \citep{maas-EtAl:2011:ACL-HLT2011} and the 20 Newsgroups dataset\footnote{\url{http://qwone.com/~jason/20Newsgroups/}} (\texttt{Newsgroups}) \citep{Lang95}. The IMDB dataset is for binary sentiment classification dataset comprising of $25,000$ highly polar positive and negative movie reviews for training and an equal number for testing. The 20 Newsgroups dataset is a collection of approximately $20,000$ newsgroup documents from 20 categories. In this study, we use a subset of $4$ categories\footnote{\texttt{comp.windows.x}, \texttt{talk.politics.misc}, \texttt{sci.space}, \texttt{talk.religion.misc}} for a four class text classification task. In total, for the 20 Newsgroups dataset, we had 2,508 training samples and 1,669 test samples.

\subsection{Weak supervision sources}
\begin{figure}[tbp!]
    \centering
    \includegraphics[width = 0.6\textwidth]{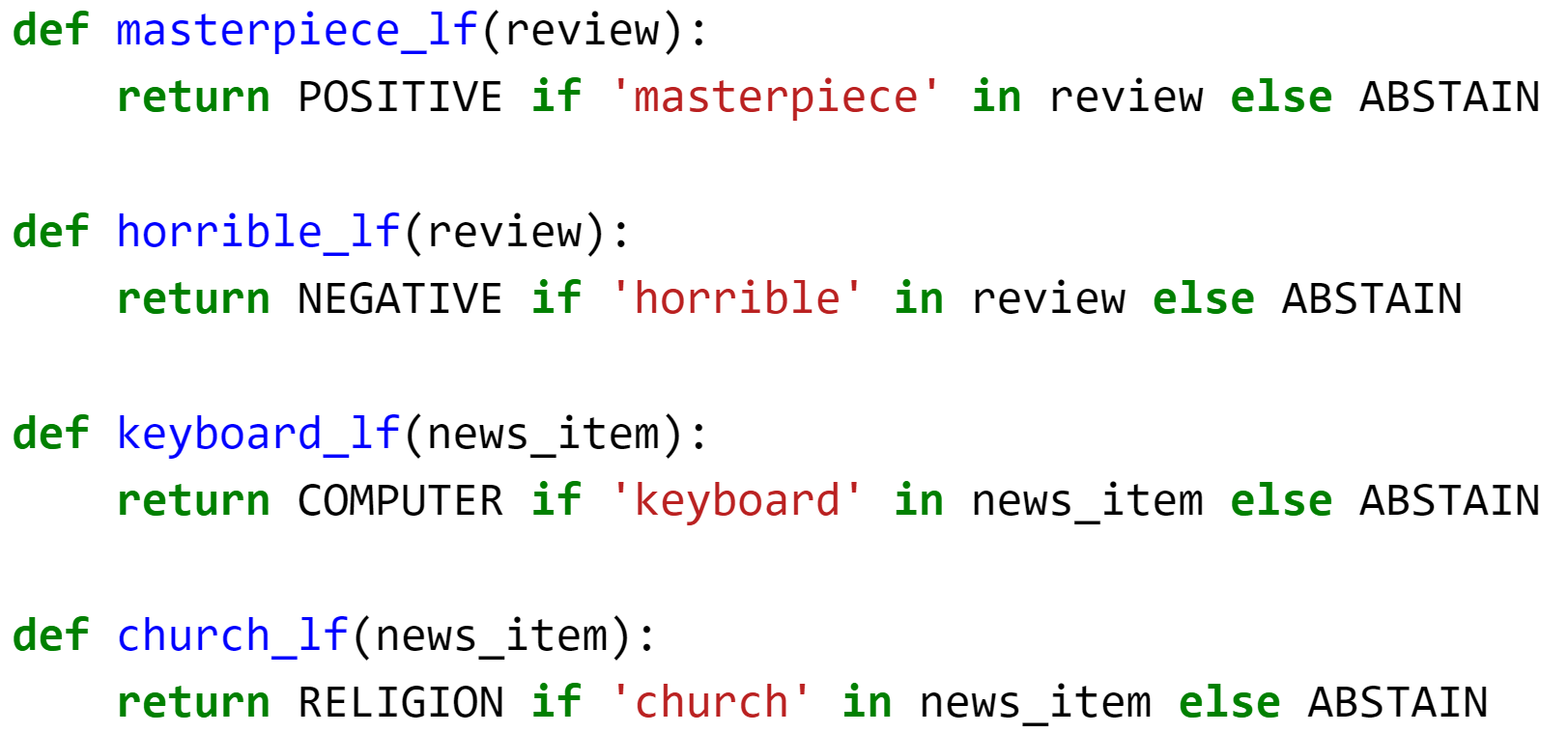}
    \caption{Examples of keyword labelling functions for both the \texttt{IMDb} and \texttt{Newsgroups} datasets.}
    \label{fig:lf_examples}
\end{figure}

For both the \texttt{IMDb} and \texttt{Newsgroups} datasets, we used simple keyword LFs as shown in Fig.~\ref{fig:lf_examples}. Keywords and key phrases have been shown to be excellent sources of weak supervision. To derive LFs we first found an exhaustive set of top $95$-percentile frequent unigrams. In the next step, we assigned these candidate keywords to categories (classes) either based on existing lexicons or similarity functions. 

For the \texttt{IMDb} dataset, we used the VADER (\textbf{V}alence \textbf{A}ware \textbf{D}ictionary and s\textbf{E}ntiment \textbf{R}easoner) lexicon \citep{gilbert2014vader} to classify the candidate keywords as being indicative of a \textit{positive}, \textit{neutral} or \textit{negative} sentiment. We finally used the positive and negative unigrams as LFs. For instance, if a review contains the keyword `\textit{masterpiece}', it is more likely to be positive. On the other hand, for the every candidate keyword in the Newsgroups training data, we assigned it to the `closest' category in terms of the cosine similarity of their corresponding BERT~\citep{devlin-etal-2019-bert} embeddings. For example, the keyword `\textit{keyboard}' is closest to `\textit{computer}' and hence suggestive of a \texttt{comp.winds.x} news item. Additionally, we choose all the words above a threshold that is similar to the category. Tab.~\ref{tab:imdbwords} and \ref{tab:newsgroupswords} list the complete set of keywords used for both the \texttt{IMDb} and \texttt{Newsgroups} dataset.

\begin{table}[!tbp]
    \resizebox{0.95\textwidth}{!}{  
    \centering
    \begin{tabular}{c|ccccc}
         \textbf{Models} & \textbf{\# labelled} & \textbf{AUC} & \textbf{Accuracy} & \textbf{FPR@50\%TPR} & \textbf{FNR@50\%TNR} \\ \hline
         Active (LC) + BERT&  1020 & 0.842 & 0.758 & 0.073 & 0.078\\
         Active (IG) + BERT&  1020 & 0.842 & 0.758 & 0.073 & 0.078\\
         Semi sup. + BERT & 1000 & 0.820 & 0.724 & 0.078 & 0.105\\
         Fully sup. + BERT &  25000 & 0.857 & 0.772 & 0.060 & 0.967\\
         Majority vote + BERT & 0 & 0.166 & 0.500 & 0.075 & 0.095\\
         Weakly sup. + BERT & \textbf{0} & \textbf{0.850} & \textbf{0.768} & \textbf{0.066} & \textbf{0.076} \\ \hline
         Fully sup. + TF-IDF &  25000 & 0.868 & 0.782 & 0.062 & 0.050\\
         Weakly sup. + TF-IDF & \textbf{0} & \textbf{0.827} & \textbf{0.749} & \textbf{0.108} & \textbf{0.076} \\ \hline
    \end{tabular}}
    \caption{Summary of results for the \texttt{IMDb} dataset. Our weakly supervised model, trained using negligible labelling effort performs at par with its fully supervised counterpart. Active (LC) and (IG) correspond to least confidence and information gain based active learning models, respectively.}
    \label{tab:IMDbresults}
\end{table}

\begin{table}[tb]
    \resizebox{0.95\textwidth}{!}{  
    \centering
    \begin{tabular}{c|ccccc}
         \textbf{Models} & \textbf{\# labelled} & \textbf{Accuracy} & \textbf{W. precision} & \textbf{W. recall} & \textbf{W. F1-score} \\ \hline
         Active (IG) + BERT &  1020 & 0.805 & 0.812 & 0.805 & 0.805\\
         Semi sup. + BERT & 1000 & 0.525 & 0.373 & 0.525 &	0.413 \\
         Fully sup. + BERT & 2028 & 0.797 & 0.799 & 0.797 & 0.797\\
         Majority vote + BERT & 0 & 0.271 & 0.142 & 0.271 & 0.128 \\
         Weakly sup. + BERT & \textbf{0} & \textbf{0.801}	& \textbf{0.802} & \textbf{0.801} & \textbf{0.800} \\ \hline
         Fully sup. + TF-IDF &  2028 & 0.772 & 0.777 & 0.772 & 0.773\\
         Weakly sup. + TF-IDF & \textbf{0} & \textbf{0.766} & \textbf{0.768} & \textbf{0.766} & \textbf{0.765}\\ \hline
    \end{tabular}}
    \caption{Summary of results for the \texttt{Newsgroups} dataset. Our weakly supervised model, trained using negligible labelling effort performs at par with its fully supervised counterpart. W. indicates weighted scores.}
    \label{tab:newgroupsresults}
\end{table}

\section{Results and Discussion}
\label{resultsanddiscussion}
Tab.~\ref{tab:IMDbresults} and~\ref{tab:newgroupsresults} summarize the results of our experiments. Some of our key findings are as follows. \\

\noindent
\textbf{Weak supervision yields competitive end models with negligible pointillistic supervision.} Our results reveal that our weakly supervised models, trained without any access to ground truth labels, performed at par with the their fully supervised counterparts, for both the \texttt{IMDb} and \texttt{Newsgroups} datasets. \\

\begin{figure*}[!bt]
    \centering
  \begin{tabular}{cccc}
    \includegraphics[width=0.25\linewidth]{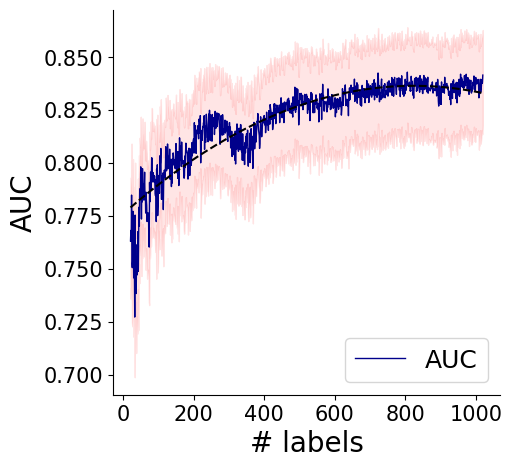} 
    &\includegraphics[width=0.25\linewidth]{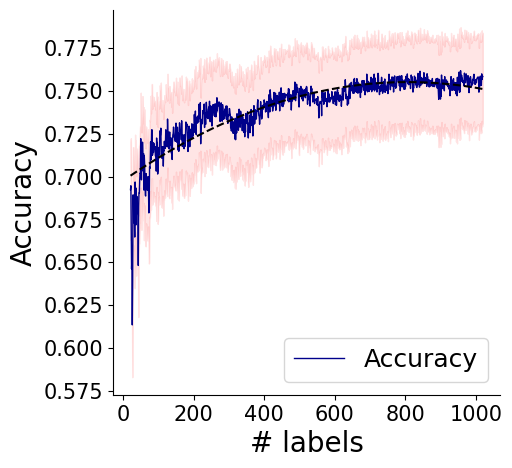}
    &\includegraphics[width=0.25\linewidth]{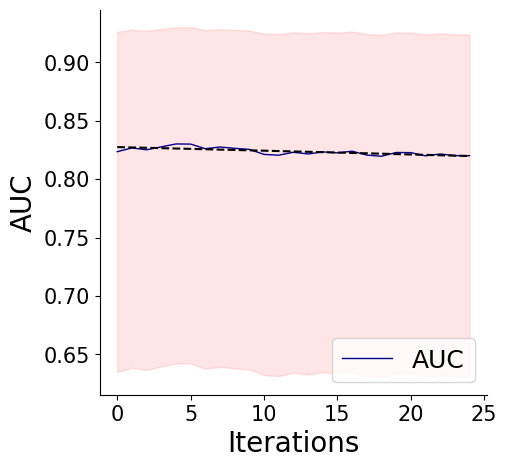} 
    &\includegraphics[width=0.25\linewidth]{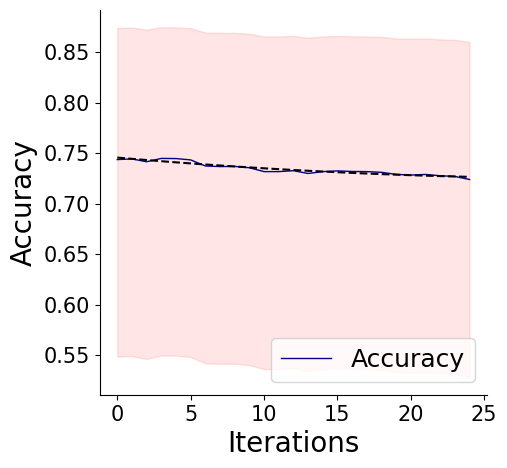} \\
    (i)&(ii)&(iii)&(iv) \\
    \includegraphics[width=0.25\linewidth]{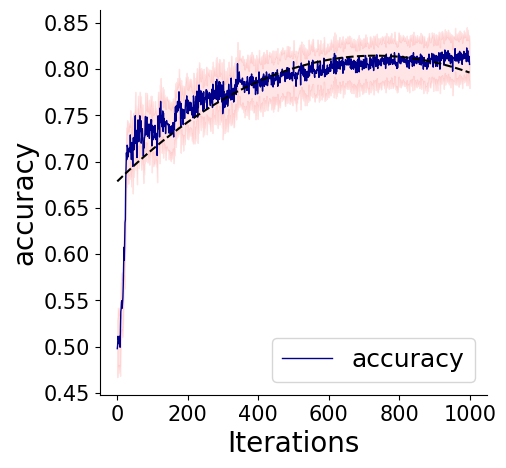} 
    &\includegraphics[width=0.25\linewidth]{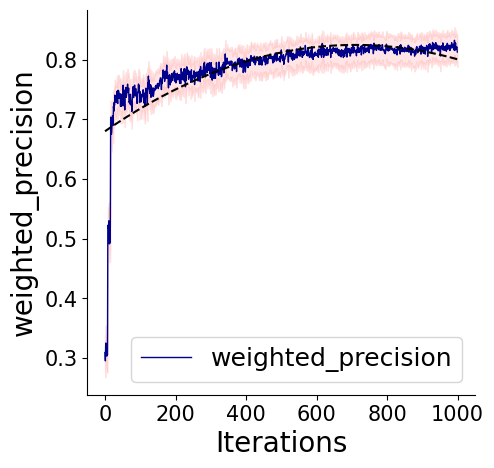}
    &\includegraphics[width=0.25\linewidth]{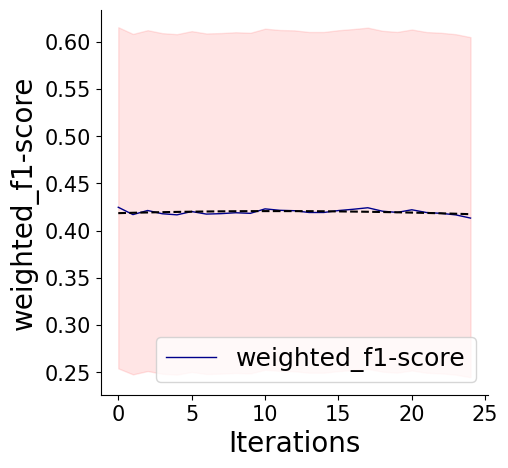} 
    &\includegraphics[width=0.25\linewidth]{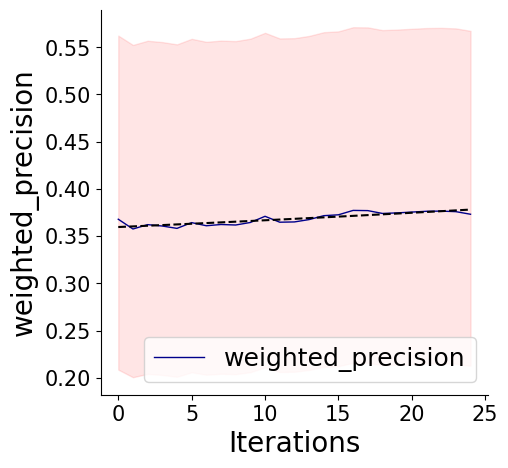} \\
    (v)&(vi)&(vii)&(viii) \\
  \end{tabular}
    \caption{Performance trajectories of information gain-based active learning (AL) (\textit{i, ii, v, vi}) and semi-supervised learning (SSL) (\textit{iii, iv, vii, viii}) with the learning iterations for the \texttt{IMDb} (i - iv) and \texttt{Newsgroups} datasets, respectively. While AL performance trajectories demonstrate a logarithmic increase, the SSL trajectories are fairly flat. The red bands are $95$\% Wilson's score confidence intervals.}
    \label{fig:learningtrajectories}
\end{figure*}

\noindent
\textbf{Alternative approaches require more effort and show inferior performance.} For both the \texttt{IMDb} and \texttt{Newsgroups} datasets, our active learning (AL) models required at least $600$ labelled samples (see Fig.~\ref{fig:learningtrajectories}) to perform at par with the supervised learning models. In fact, in the \texttt{Newgroups} dataset, our AL model outperforms both the full and weakly supervised models. Our findings are consistent with several studies ~\citep{settles2009active} which have theoretically shown that AL can achieve the same error rate as full supervision with fewer but intelligently chosen query objects. In comparison, we found that self-labelling semi-supervised learning (SSL) did not demonstrate a significant improvement in performance over the performance in the initial seed set. In fact, in the \texttt{IMDb} dataset, we observed a degradation in performance as the end model classifiers were re-trained on more self-labelled data points (see Fig.~\ref{fig:learningtrajectories}). \\

\noindent
\textbf{Weak supervision significantly outperformed majority voting.} Recall that while majority voting weighs each LF equally, data programming weighed each LF in proportion to its empirical accuracy estimated by the label model (see Eqns.~\ref{eqn:weightedmajorityvoting} and \ref{eqn:probabilisticlabels}). Our results illustrate the distinct advantage of the accuracy-weighted majority voting over the simple majority voting. \\

\noindent
\textbf{BERT does not offer a significant advantage over TF-IDF embeddings.} In addition to using BERT embeddings, we also compared the performance of our weakly and fully supervised models using TF-IDF (Term Frequency, Inverse Document Frequency) embeddings as featurization. Since TF-IDF embeddings are sparse\footnote{They are of the same length as the vocabulary}, instead of using the raw embeddings, we used the first $500$ components obtained from the Principal Components Analysis (PCA) algorithm. We found that both the weakly and full supervised models demonstrated comparable performance in both the settings. However, we observed a natural deterioration of performance when using TF-IDF instead of BERT embeddings. This is an important result since BERT is pre-trained on a very large corpus and offers another source of weak supervision. The goal of this experiment was to quantify if this provided any advantages to the weakly supervised model. \\

\noindent
\textbf{Keywords are excellent sources of weak supervision.} The major finding of our study is that keywords are excellent sources of weak supervision for the data programming label model. While we were able to easily and automatically ``gather'' supervision at the unigram level using existing lexicons and similarity functions, for more involved problems, such alternate supervision sources may be hard to find. Moreover, the limitations of our automated keyword based methodology may simply not be visible given the nature of the datasets we considered. However, even in this case, we may still be able to efficiently train a simple keyword classifier using active learning. We must note that obtaining supervision at the keyword or keyphrase level is still much more efficient than labelling the entire corpus since the manual effort is upper-bounded by the size of the \textit{frequent} vocabulary\footnote{Top \textit{n}\% frequent words} which is usually  significantly smaller than size of the corpus.

\section{Conclusion}
In this work we show that weak supervision is a competitive alternative to fully supervised classification models. Our results over multiple text classification datasets reveal that weak supervision performs at par with its fully supervised counterpart without access to pointillistic ground truth labels. We found that the considered alternatives--majority voting, active, and semi-supervised learning demonstrated inferior performance despite relying on more manual effort. Our results also underscored the importance of the accuracy-weighted majority voting aggregation followed by data programming in comparison to simple majority voting. Future work involves empirically evaluating the performance of keywords and keyphrases as sources of weak supervision over multiple datasets as well as extensions to applications beyond text data.


\bibliography{bibliography}

\appendix
\section{Keyword LFs}\label{apd:first}

\begin{table}[h!]
\resizebox{0.95\textwidth}{!}{  
\begin{tabular}{l|l}
\multicolumn{1}{c}{\textbf{Sentiment}} & \multicolumn{1}{|c}{\textbf{Words}}\\ \hline
Positive                        & \begin{tabular}[c]{@{}l@{}}'amazing', 'amusing', 'appreciate', 'attractive', 'awesome', 'beautiful', 'beautifully', 'best', \\ 'better', 'brilliant', 'certainly', 'charm', 'charming', 'clearly', 'clever', 'compelling', \\ 'convincing', 'cool', 'creative', 'definitely', 'delightful', 'effective', 'engaging', \\ 'enjoy', 'enjoyable', 'enjoyed', 'entertaining', 'entertainment', 'excellent', 'exciting', 'fantastic', \\ 'fascinating', 'favorite', 'fine', 'fun', 'funniest', 'funny', 'glad', 'good', \\ 'gorgeous', 'great', 'greatest', 'happy', 'hilarious', 'impressed', 'impression', 'impressive', \\ 'interesting', 'laugh', 'laughed', 'laughing', 'laughs', 'like', 'liked', 'likes', 'love', \\ 'loved', 'lovely', 'loves', 'loving', 'lucky', 'masterpiece', 'nice', 'nicely', 'novel', 'original', \\ 'outstanding', 'perfect', 'perfectly', 'pleasure', 'powerful', 'recommend', 'recommended', 'remarkable', \\ 'smart', 'solid', 'special', 'strong', 'stunning', 'super', 'superb', 'superior', 'surprise', 'surprised',\\ 'surprising', 'surprisingly', 'terrific', 'top', 'treat', 'truly', 'unbelievable', 'well', 'wonderful', \\ 'wonderfully', 'worth', 'worthy', 'wow'\end{tabular} \\ \hline
Negative                        & \begin{tabular}[c]{@{}l@{}}'avoid', 'awful', 'bad', 'badly', 'bizarre', 'bored', 'boring', 'bother', 'confused', 'confusing', 'crap',\\ 'desperate', 'disappointed', 'disappointing', 'disappointment', 'dull', 'dumb', 'fail', 'failed', 'fails',  \\ 'forced', 'forget', 'forgotten',  'hard', 'hate', 'hated', 'hell', 'horrible', 'ill',  'lack', 'lame',  \\ 'mess', 'miss', 'missed', 'missing', 'mistake', 'nasty', 'negative', 'nonsense', 'odd', 'pain', 'painful', \\ 'pathetic', 'poor', 'ridiculous', 'sad', 'sadly', 'shame', 'sick', 'stupid', \\ 'sucks', 'terrible', 'terribly', 'tired', 'torture', 'ugly', 'unfortunately', 'warning', 'waste', \\ 'wasted', 'weak', 'weird', 'worse', 'worst', 'wrong', 'not', 'don\textbackslash{}'t'\end{tabular}\\ \hline
\end{tabular}
}
\caption{Positive and negative keywords for \texttt{IMDb} dataset}
\label{tab:imdbwords}
\end{table}
\begin{table}[h!]
\resizebox{0.95\textwidth}{!}{  
\begin{tabular}{l|l}
\multicolumn{1}{c|}{\textbf{Category}} & \multicolumn{1}{c}{\textbf{Words}}\\ \hline
religion                       & \begin{tabular}[c]{@{}l@{}}'religion', 'religions', 'religious', 'christianity', 'christians', 'church', 'faith', \\ 'christian', 'beliefs', 'holy', 'christ', 'bible', 'belief', 'god', 'jesus', 'islam', 'mass', \\ 'mary', 'biblical', 'cross', 'matthew', 'moral', 'convert', 'islamic', 'paul', 'followers', \\ 'country', 'john', 'values', 'believe', 'community', 'believing', 'morality', 'message', 'lord', \\ 'nature', 'sites', 'peter'\end{tabular}                                                                   \\ \hline
politics                       & \begin{tabular}[c]{@{}l@{}}'political', 'debate', 'opinions', 'government', 'opinion', 'discussion', 'policy', \\ 'discuss', 'congress', 'talk', 'topic', 'arguments', 'history', 'argument', 'dc', 'report', \\ 'media', 'leaders', 'reports', 'business', 'issue', 'issues', 'law', 'sources', 'president', \\ 'police', 'comments', 'comment', 'document', 'phone', 'companies', 'statements', 'meeting', \\ 'news', 'articles', 'events', 'washington', 'war',  'official', 'military', 'court'\end{tabular}                   \\ \hline
computer                       & \begin{tabular}[c]{@{}l@{}}'pc', 'hardware', 'software', 'keyboard', 'machine', 'machines', 'mouse', 'unix', \\ 'graphics', 'ibm', 'mode', 'interface', 'internet', 'mac', 'user', 'screen', 'systems', 'compile', \\ 'file', 'com', 'technology', 'files', 'usenet', 'memory', 'input', 'widget', 'programming', 'system', \\ 'board', 'comp', 'function', 'technical', 'info', 'output',  'levels', 'functions', \\ 'button', 'display', 'engineering', 'control', 'widgets', 'library', 'code', 'os', \\ 'program'\end{tabular} \\ \hline
space                          & \begin{tabular}[c]{@{}l@{}}'mars', 'spacecraft', 'orbit', 'nasa', 'orbital', 'planet', 'universe', 'planetary', 'rocket',\\ 'moon', 'satellites', 'shuttle', 'rockets', 'satellite', 'apollo', 'flight', 'plane', 'physics', 'launch',\\ 'fly', 'air', 'mission', 'astronomy', 'nuclear', 'solar', 'earth', 'world', 'configuration', 'lunar', 'energy'\end{tabular}\\ \hline
\end{tabular}
}
\caption{Category-related keywords for the \texttt{Newsgroups} dataset}
\label{tab:newsgroupswords}
\end{table}

\end{document}